\documentclass[3p,times,procedia]{elsarticle}
\usepackage{ecrc}
\usepackage{csquotes}
\usepackage{booktabs}

\usepackage[bookmarks=false]{hyperref}
    \hypersetup{colorlinks,
      linkcolor=blue,
      citecolor=blue,
      urlcolor=blue}

\volume{00}

\firstpage{1}

\journalname{Procedia Computer Science}

\runauth{Author name}

\jid{procs}

\usepackage{amssymb}

\usepackage[figuresright]{rotating}

\begin{document}
\begin{frontmatter}



\dochead{29th International Conference on Knowledge-Based and Intelligent Information \& Engineering Systems (KES 2025)}%

\title{Benchmarking Large Language Models for Biomedical Relation Extraction}

\author[a,c]{Claudiu Creangă} 
\author[b,c]{Teodor Marchitan}
\author[b,c]{Liviu P. Dinu}

\address[a]{Interdisciplinary School of Doctoral Studies}
\address[b]{Faculty of Mathematics and Computer Science, University of Bucharest, Romania}
\address[c]{HLT Research Center}

\begin{abstract}
Extracting SNP-phenotype associations from biomedical literature is vital but challenging. We benchmarked diverse NLP models, including MLMs, hybrid architectures, and state-of-the-art LLMs (Gemini 2.0, OpenAI O-series, Qwen, Mistral), on the SNPPhenA corpus across three tasks: sentence-level, abstract-level, and association strength classification. OpenAI O1 achieved state-of-the-art (SOTA) results using few-shot learning for non-finetuned sentence-level classification (F1 0.89) and established a new SOTA for abstract-level classification (F1 0.82). Association strength classification proved difficult, though fine-tuned Gemini 2.0 Pro performed best (F1 0.60) in the first LLM evaluation of this task. Proprietary LLMs, especially in few-shot (O1) or fine-tuned (Gemini 2.0 Pro) settings, significantly outperformed other models. These findings confirm the power of modern LLMs for genomic knowledge extraction.
\end{abstract}


\end{frontmatter}

\email{claudiu.creanga@fmi.unibuc.ro}



\section{Introduction}
\label{introduction}

Identifying the connections between Single Nucleotide Polymorphisms (SNPs) and phenotypes is an important area of research in computational biology and biomedical informatics \cite{Satam2023}. This specialized form of relationship extraction bridges multiple domains, transforming unstructured textual data into actionable genetic insights that can revolutionize our understanding of human health and disease. Scientific literature on genetic associations has grown exponentially, making manual curation of genetic relationships virtually impossible. Natural Language Processing (NLP) techniques offer a scalable solution to systematically extract and organize complex genetic relationships from vast amounts of textual data. By automating the identification of SNP-phenotype connections, researchers can rapidly synthesize knowledge that would take human experts years to compile.

Extracting relationships between SNPs and phenotypes faces several difficulties. Genetic literature uses specialized terminology, and these relationships are often described using complex language that depends on the context. Understanding the subtle meanings requires careful natural language processing, as genetic variations can have different interpretations depending on the situation. This complexity means that simple methods like pattern matching or keyword extraction are often insufficient. The research landscape for SNP-phenotype relationship extraction remains relatively unexplored within NLP, presenting a significant opportunity for groundbreaking work. Therefore, improvements in NLP techniques could help us extract more useful information from genetic literature. Developing better methods to understand the specific language used in biology is important for identifying and interpreting these complex relationships. This work could lead to a better understanding of the genetic basis of different traits and diseases.

\subsection{Contribution}
In this paper we take the SNPPhenA corpus \cite{Bokharaeian2017} and focus on 3 tasks: 

\begin{itemize}
    \item \textbf{Sentence-Level SNP-Phenotype Association Classification:} Given an individual sentence from the SNPPhenA corpus, the task is to predict the nature of the association between a specific SNP and a phenotype. The possible association categories are \textit{positive}, \textit{negative}, or \textit{neutral}. We achieve new SOTA for models that are not fine-tuned for the task. 
    \item \textbf{Abstract-Level SNP-Phenotype Association Classification:} For this task, the input is an entire medical abstract from the SNPPhenA corpus. The objective is to classify the overall association between a SNP and a phenotype as either \textit{positive}, \textit{negative}, or \textit{neutral}, based on the information presented in the whole abstract. We achieve new SOTA with the latest models. 
    \item \textbf{SNP-Phenotype Association Strength Classification:} This task focuses on predicting the strength of the association between a SNP and a phenotype. Given  a sentence which has a positive association between a SNP and a phenotype, the model must classify the association as having a \textit{weak}, \textit{moderate}, or \textit{strong} relationship. This is the first study to test LLMs on this task. 
\end{itemize}

\section{Related Work}
\label{related}
Building upon the foundational work of our previous study \cite{Creanga2024}, this paper aims to further explore the landscape of language models for biomedical relation extraction. Our prior research culminated in demonstrating the exceptional capabilities of a fine-tuned Gemini 1.0 Pro model when applied to the task of identifying relationships between SNPs and phenotypes within the SNPPhenA corpus. This investigation revealed that a fine-tuned Gemini 1.0 Pro surpassed existing state-of-the-art results on this dataset. Notably, this achievement was observed not only at the sentence level, where we attained an F1-score of 0.89, outperforming previous models like BioBERT-GRU \cite{Dehghani2023} and PubMedBERT-LSTM, but also at the more complex abstract level, where we achieved an F1-score of 0.80, significantly improving upon the prior best of 0.645 held by BioBERTGRU. These findings underscored the potential of large language models, particularly those with the architecture and scale of Gemini, to revolutionize automated knowledge extraction from the vast and ever-growing body of biomedical literature. 

While the performance of fine-tuned Gemini 1.0 Pro marked a significant advancement, the rapid evolution of LLMs necessitates a continuous exploration of newer architectures and more efficient implementations. Furthermore, the emergence of powerful open source LLMs which can be run locally, such as Qwen, Mistral or Deepseek presents an exciting avenue for research. Evaluating these open-source alternatives is important for enabling accessibility within the research community, potentially lowering the barrier to entry for institutions with limited cloud computing resources. By comparing the performance of both proprietary models (like Gemini or ChatGPT) and open-source options (like Qwen and Mistral), we aim to provide a comprehensive understanding of the trade-offs involved, ultimately guiding researchers and practitioners towards the most effective and accessible solutions for biomedical relation extraction.

\section{Dataset}

To facilitate the automatic extraction of the associations between SNPs and their resulting phenotype from biomedical literature, several corpora have been developed. Among these, the SNPPhenA corpus stands out as a valuable resource. This corpus addresses the limitations of earlier datasets by incorporating annotations for linguistic-based negation, modality markers, neutral candidates, and the confidence level of the identified SNP-phenotype associations.

These associations were manually classified into three categories: positive, negative, and neutral. Positive associations indicate a clear link between the SNP and phenotype, while negative associations denote the absence of such a link. The inclusion of a neutral category is particularly noteworthy, as it captures instances where an association is neither explicitly stated nor confidently inferred. Furthermore, for positive associations, the corpus provides a graded confidence level – weak, moderate, or strong – based on the linguistic cues within the text, including modality markers, adverbs, and reported statistical significance.

\begin{figure}[t]\vspace*{3pt}
\centerline{\includegraphics[width=1\linewidth]{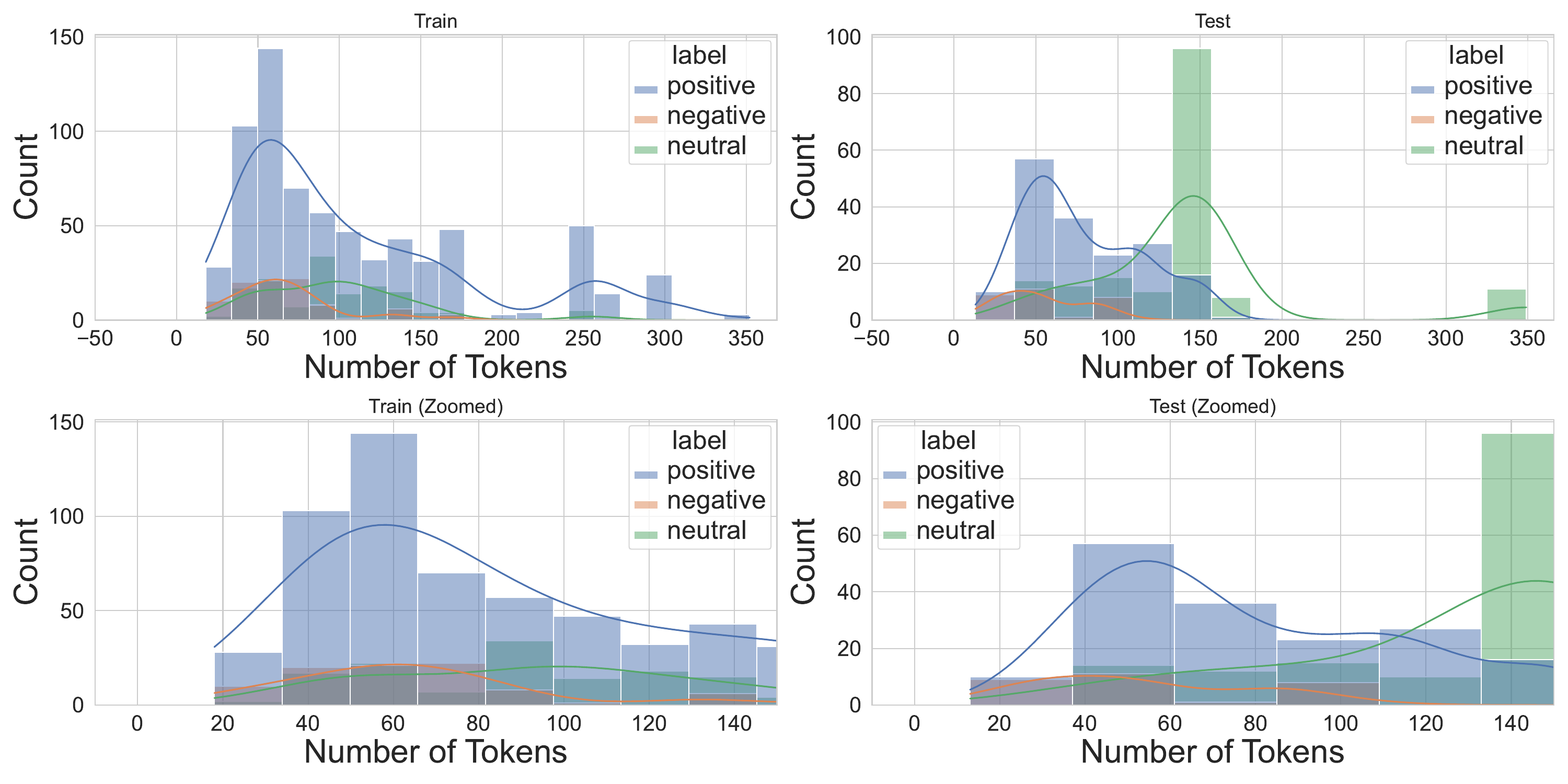}}
\caption{Distribution of the number of tokens per sentence in the SNPPhenA corpus, separated by training and testing sets and colored by the association label (positive, negative, neutral). The zoomed-in views provide a detailed perspective on sentences with lower token counts.}
\label{fig:dataset}
\end{figure}

The corpus comprises 360 XML files containing 2625 sentences, with 483 sentences annotated as containing at least one SNP and one phenotype. A total of 875 SNP names and a diverse range of phenotype terms are annotated within the corpus. The  \autoref{fig:dataset} illustrates the distribution of the number of tokens within sentences of the SNPPhenA corpus, separated into training and testing sets and further categorized by the association label: positive, negative, and neutral.

In the training set, the distribution of the total number of tokens appears to be right-skewed, with a majority of sentences containing between approximately 20 and 100 tokens. There are noticeable peaks around the 50-token mark and another around 150 tokens, with a smaller peak near 250 tokens. When examining the distribution by label in the training set, positive associations seem to be most prevalent in the range of 50 to 100 tokens, while negative and neutral associations also show a similar trend but with fewer instances overall. The zoomed-in view of the training set for sentences with fewer tokens (0-140) confirms that the majority of instances across all three labels fall within this range, with positive associations having the highest count. The testing set counts are generally lower than in the training set. Notably, the distribution for neutral associations in the testing set seems to have a more pronounced presence in the higher token count range (around 150 tokens) compared to the training set. The comparison between training and testing sets reveals important disparities in token count distributions and label proportions. While the dataset is predominantly composed of moderate-length sentences, there's an intriguing asymmetry in label representation: the training set exhibits a higher frequency of positively associated SNP-phenotype mentions, whereas the testing set shows a more pronounced proportion of \textbf{neutral associations}, especially within the 100-200 token range. This strategic data partitioning suggests a deliberate approach to model development—likely designed to challenge machine learning models to \textbf{generalize} more effectively by presenting a testing scenario that differs from the training distribution. Such a methodology can help assess a model's robustness and ability to perform consistently across varied data landscapes.

We selected the SNPPhenA corpus \cite{Bokharaeian2017} as our primary dataset due to its unique position as the sole source of comprehensively annotated, ranked SNP-phenotype associations. Despite its high-quality manual curation by three domain experts, the SNPPhenA corpus remains underutilized, having been referenced in only four published studies to date: (\cite{Bokharaeian2017}, \cite{Dehghani2023}, \cite{Bokharaeian2023} and \cite{Creanga2024}).
To contextualize the variant-phenotype relation extraction task, we provide an illustrative abstract from a biomedical text:

\begin{displayquote}
LD (r2=0.35) between \textbf{TOMM40 (rs2075650)} and \textbf{APOC1 (rs1064725)} was observed in \textbf{PPA}, but not in controls and in \textbf{bvFTD}. Inside this region of 26.9 kb, LD (r2 \textgreater 0.50) between TOMM40 (rs2075650) and APOE (rs429358) was observed in bvFTD and in controls, but not in PPA. Inside this region of 16.3 kb, LD (r2 = 0.14) between TOMM40 (rs157590) and APOE (rs429358) was observed in PPA, but not in bvFTD and in controls (emphasis mine) \cite{Seripa2012}.
\end{displayquote}

\begin{figure}[t]\vspace*{3pt}
\centerline{\includegraphics[width=1\linewidth]{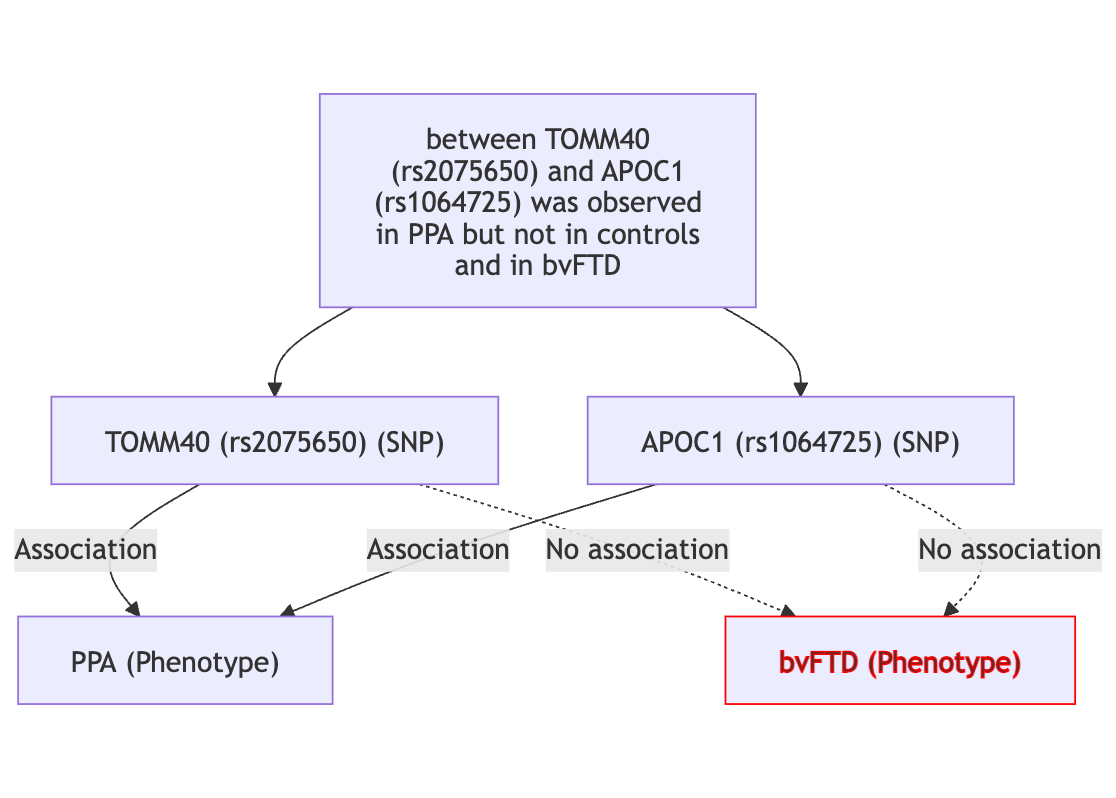}}
\caption{ A sentence from a medical abstract. This sentence illustrates the association of both SNPs TOMM40 (rs2075650) and APOC1 (rs1064725) with the PPA but not with the bvFTD phenotypes. }
\label{fig:association}
\end{figure}

We can observe three sentences that mention a relationship between one or two SNPs and a phenotype. In that abstract, the SNPs mentioned are rs1064725, rs429358, rs157590, and rs429358, and the phenotypes mentioned are PPA (Primary progressive aphasia) and bvFTD (Behavioral variant frontotemporal dementia). These entities are extracted using an NER tool. Based on the provided abstract, our REL model builds a system that detects the relation as in \autoref{fig:association}.

\section{System Overview}

We evaluated three types of systems: the largest currently available LLMs, the largest available Masked Language Models, and a hybrid deep learning model architecture.

The current SOTA results for the SNPPhenA corpus at sentence-level are shown in \autoref{table:SNPPhrenA}. 

\begin{table}[h]
\caption{SOTA for SNPPhenA corpus (sentence level).}
\begin{tabular*}{\hsize}{@{\extracolsep{\fill}}llll@{}}
\toprule
Model & Precision & Recall & F1 score (macro) \\
\colrule
DeBERTa v3 & 0.85 & 0.86 & 0.85 \\
PubMedBERT-LSTM & 0.870 & 0.883 & 0.866\\
BioBERTGRU & 0.883 & 0.882 & 0.881 \\
\textbf{Gemini 1.0 Pro Fine-tuned} & \textbf{0.88} & \textbf{0.90} & \textbf{0.89} \\
\botrule
\label{table:SNPPhrenA}
\end{tabular*}
\end{table}

The current SOTA results for the SNPPhenA corpus at abstract-level are shown in \autoref{table:SNPPhrenA_abstract}. 

\begin{table}[h]
\caption{SOTA for SNPPhenA corpus (abstract level).}
\begin{tabular*}{\hsize}{@{\extracolsep{\fill}}llll@{}}
\toprule
Model & Environment & F1 score (macro) \\
\colrule
BioBERTGRU & - & 0.645 \\
Gemini 1.5 Pro & Few-shot, CoT & 0.73 \\
\textbf{Gemini 1.0 Pro} & \textbf{Fine-tuned - balanced train dataset, CoT} & \textbf{0.80} \\
\botrule
\label{table:SNPPhrenA_abstract}
\end{tabular*}
\end{table}

The current SOTA results for the SNPPhenA corpus for determining the strength of associations is 0.45. The only attempt that we are aware of is the one presented in \cite{Bokharaeian2017}, which is a BoW based model..

\subsection{Masked Language Models}

Despite advancements in other architectures, finetuning BERT-based models continues to achieve SOTA results in various classification tasks, like NER, so it is still worthwhile to try them. MLMs learn by strategically masking words within a sentence and predicting the missing words based on the surrounding bi-directional context. This approach differs from causal models, which are autoregressive and predict the next token based solely on preceding tokens. We experimented with DeBERTa V3 \cite{He2020}, BERT large \cite{Devlin2018}, XLM-RoBERTa \cite{Conneau2019}, and ModernBERT \cite{Warner2024}, the latest large-scale BERT model. 

We augment the pre-trained model with a custom output layer that includes several  components: first, a normalization layer to stabilize inputs, followed by a hyperbolic tangent (tanh) activation function to introduce non-linearity. We then apply a dropout layer to prevent overfitting, and conclude with a linear layer that reduces the output to a single dimension for binary classification.

\begin{figure}[t]\vspace*{3pt}
\centerline{\includegraphics[width=1\linewidth]{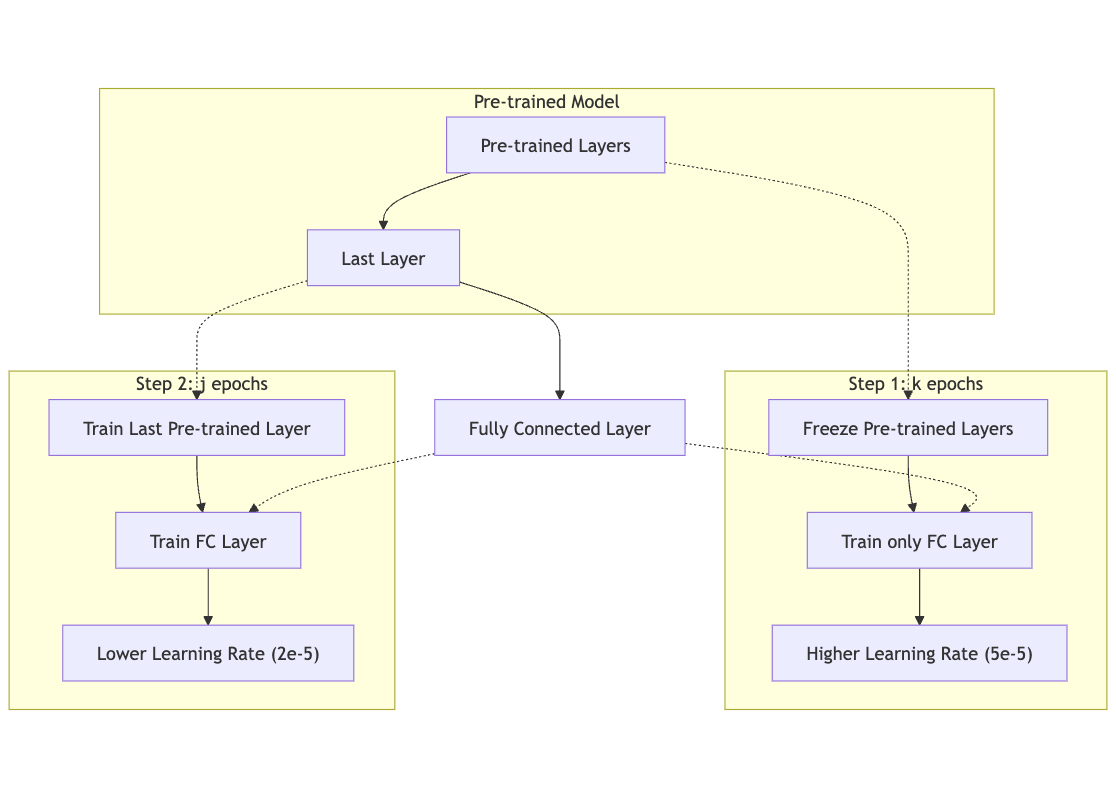}}
\caption{ A two-step fine-tuning approach: Step 1 involves training only the newly added fully connected layer with a higher learning rate, while the pre-trained layers are frozen. Step 2 then trains both the last pre-trained layer and the fully connected layer with a lower learning rate.}
\label{fig:finetuning}
\end{figure}

Our fine-tuning methodology leverages a sequential two-stage training regimen for pre-trained transformer models \autoref{fig:finetuning}. During the initial k epochs, we enforce complete weight freezing across all transformer layers, effectively preserving the pre-trained representations. Subsequently, for the remaining $N{-}k$ epochs, we selectively enable gradient updates only for the designated feature extraction layers. This strategy aims to optimize resource allocation while retaining the generalizable knowledge embedded within the lower transformer layers and focusing adaptation on task-relevant features extracted from the upper layers. The determination of optimal epoch counts for both stages was guided by monitoring the validation loss and test set performance to prevent overfitting. Feature extraction was performed by taking the output of the [CLS] token from the last hidden layer, followed by a dropout rate of 0.5. A fully connected classifier, comprising two linear layers with output dimensions [256,64], an intermediate dropout of 0.5, and a final sigmoid activation, was employed for prediction. A specific training run involved 2 epochs with the entire transformer frozen, a batch size of 64, the AdamW optimizer with a learning rate of 2e-4, and the binary cross-entropy loss function. Our primary layer selection strategy involved utilizing the last transformer layer. The best performing model was ModernBERT, as see in in \autoref{table:sentence_level}.

\subsection{Hybrid Architecture}

\begin{figure}[t]\vspace*{3pt}
\centerline{\includegraphics[width=0.8\linewidth]{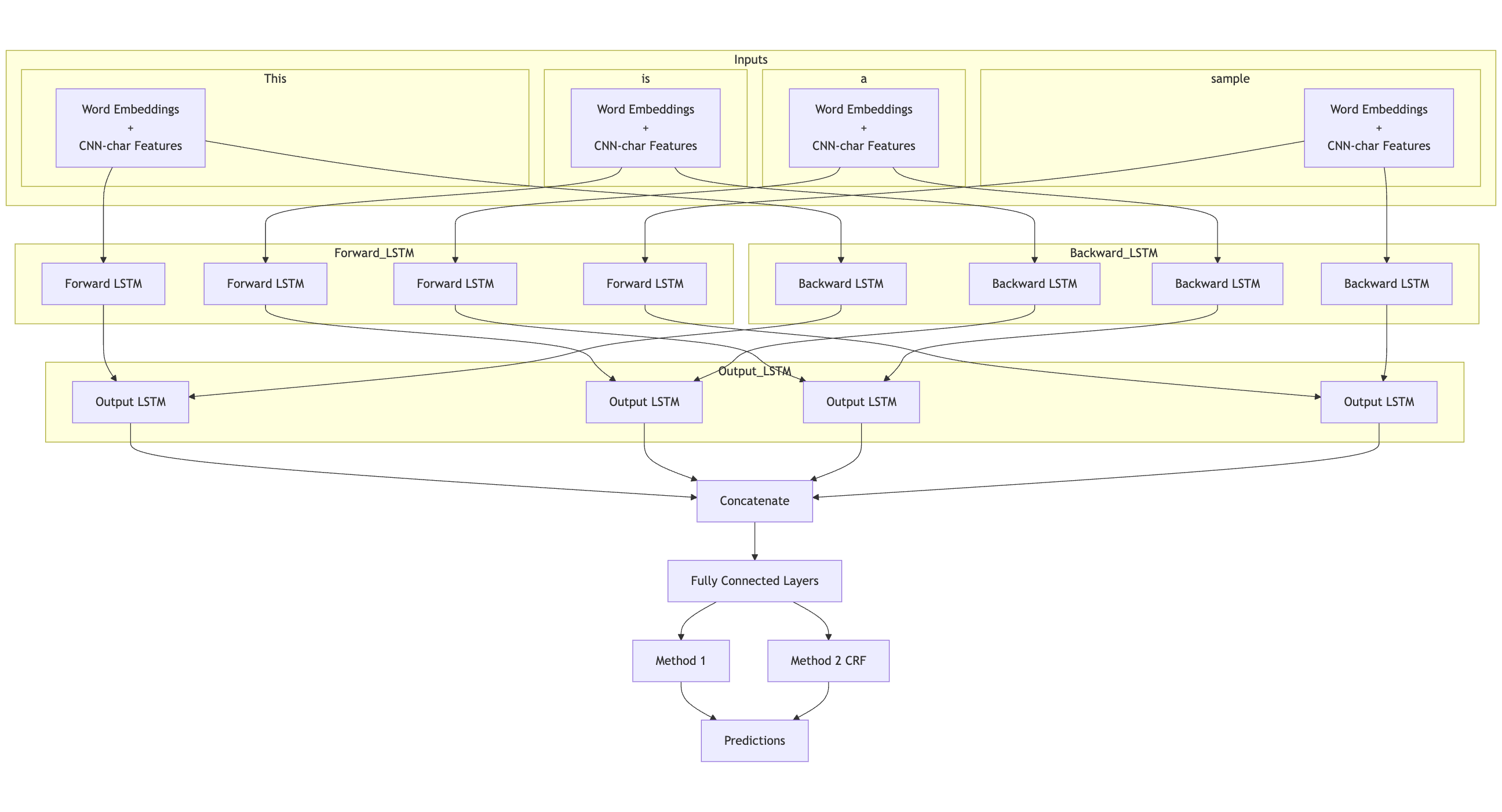}}
\caption{Hybrid neural network architecture combining character-level CNN, word embeddings, bidirectional LSTM, and an optional CRF layer.}
\label{fig:hybrid}
\end{figure}

Our model architecture follows the approach proposed by \cite{Chiu2016}, which demonstrated significant effectiveness in named entity recognition tasks. Our implementation is described in \autoref{fig:hybrid}. The method leverages a hybrid approach to input representation by transforming words and characters into vector embeddings that can be processed by a neural network. The feature extraction begins with character-level representations, where we utilize a lookup table for character embeddings and apply a one-dimensional CNN with max pooling. This approach allows the model to capture intricate morphological features of individual characters. Simultaneously, word-level features are generated through word embeddings created using a separate lookup table. These character and word-level features are concatenated to create a rich, comprehensive input representation. The combined features are then processed through a bidirectional LSTM network, which enables the model to capture contextual dependencies in both forward and backward directions. The bidirectional LSTM outputs are subsequently passed through fully connected layers, with the architecture supporting multiple prediction methods. An optional Conditional Random Field (CRF) layer can be incorporated to enhance sequence labeling performance by modeling interdependencies between output labels. 

Despite the architectural sophistication, our experimental results revealed that this hybrid neural network approach underperformed relative to contemporary LLMs and MLMs, as seen in \autoref{table:sentence_level}.

\subsection{Causal Models}

We evaluated the performance of several open-source and proprietary causal models.

For open-source causal models, we assessed Mistral-7B-Instruct-v0.2 \cite{jiang2023mistral}, Qwen2.5:14b \cite{Bai2023}, Qwen2.5:32b-instruct-q2, and Deepseek-r1:14b-qwen-distill.

Among the proprietary models, we evaluated Gemini 2.0 Flash, Gemini 2.0 Pro, Gemini 2.0 Flash Thinking, OpenAI GPT-4-Turbo, OpenAI GPT-4o, OpenAI o1-mini, OpenAI o3-mini, and OpenAI o1.

These models were tested in both zero-shot and few-shot settings. Furthermore, some models underwent fine-tuning on the training dataset.

In the zero-shot setting, the models were provided with a sample for testing and rules for predicting the class, without any prior training data or examples.

The following prompt yielded the best results:

\begin{displayquote}
Given a sentence, predict the association between the SNP and the phenotype. 
Chose one of the following labels: positive, neutral, negative. 
Chose positive only if there is a clear indicated association. 
Chose negative only if a lack of association is evident.
Otherwise, for those that fall between the two other classes, where the presence or absence of association is not remarked in the sentence (or it is only potential), choose neutral.

This is the sentence to predict: $[sentence]$
The SNP: $[snp]$
The Phenotype: $[phenotype]$
Output:

Return only the association between the SNP: $[snp]$ and Phenotype: $[phenotype]$, which is one of the 3 values (positive, neutral, negative), with no additional text, no explanations, or formatting. No code at all.                
\end{displayquote}

In the few-shot setting, the models were provided with one or more examples from each class within the same prompt to guide them on how to solve the task. Model fine-tuning was performed using a subset of the training set. This subset comprised a balanced collection of 50 examples from each class. We minimized variability by setting the LLM temperature to 0 and running each test three times.

\section{Results}

\subsection{Performance at first task: sentence level}

From \autoref{table:sentence_level} we can see that OpenAI O1 achieves the highest performance on the sentence-level task, which is SOTA for a non-finetuned model, closely followed by fine-tuned Gemini 2.0 Pro. Interestingly, fine-tuning Gemini 2.0 Flash performed worst than few shot techniques. Open source models like Qwen2.5, Mistral-7B, and Deepseek-r1 Qwen Distill demonstrate comparatively lower effectiveness on this task. 

\begin{table}[h]
\caption{Performance evaluation of multiple AI models on the sentence-level task, showing Precision, Recall, and F1 scores across different settings, with OpenAI O1 achieving the highest F1 score in a few-shot environment.}
\begin{tabular*}{\hsize}{@{\extracolsep{\fill}}lllll@{}}
\toprule
Model & Environment & Precision & Recall & F1 score (macro) \\
\colrule
DeBERTa V3 Hybrid & fine-tune & 0.56 & 0.50 & 0.52 \\
XLM-RoBERTa & fine-tune & 0.54 & 0.54 & 0.55 \\
DeBERTa V3 & fine-tune & 0.54 & 0.57 & 0.56 \\
XLM-RoBERTa Hybrid & fine-tune & 0.60 & 0.58 & 0.58 \\
Deepseek-r1:14b-qwen-distill & few-shot 1 example & 0.61 & 0.63 & 0.62 \\
ModernBERT & fine-tune & 0.64 & 0.64 & 0.63 \\
Mistral-7B-Instruct-v0.2 & few-shot 1 example & 0.66 & 0.67 & 0.66 \\
Qwen2.5:32b-q2-K & few-shot 1 example & 0.74 & 0.75 & 0.70 \\
Qwen2.5:14b & few-shot 1 example & 0.77 & 0.76 & 0.74 \\
Gemini 2.0 Flash & fine-tune & 0.77 & 0.83 & 0.78 \\
Gemini 2.0 Flash & few shot 50 examples & 0.77 & 0.81 & 0.79 \\
OpenAI GPT-4o & few-shot 1 example & 0.83 & 0.80 & 0.79 \\
Gemini 2.0 Pro & zero shot & 0.79 & 0.85 & 0.81 \\
Gemini 2.0 Flash Thinking & few-shot 1 example & 0.85 & 0.86 & 0.85 \\
Gemini 2.0 Pro & few-shot 1 example & 0.87 & 0.86 & 0.86 \\
OpenAI O3-mini & few-shot 1 example & 0.87 & 0.87 & 0.87 \\
Gemini 2.0 Pro & fine-tune & 0.84 & \textbf{0.90} & 0.87 \\
\textbf{OpenAI O1} & \textbf{few-shot 1 example} & \textbf{0.89} & 0.89 & \textbf{0.89} \\
\bottomrule
\label{table:sentence_level}
\end{tabular*}
\end{table}

\subsection{Performance at second task: abstract level}
We experimented with the best models from each family from the previous task and we can see in  \autoref{table:abstract_level} that the performance of causal models varied significantly, with OpenAI O1 emerging as the top performer by achieving the highest scores across metrics, while Qwen exhibited the lowest performance. With OpenAI O1 model we obtained new SOTA at predicting SNP-phenotype associations from medical abstracts. 

\begin{table}[h]
\caption{Comparison of our best models (Precision, Recall, F1) for the abstract-level task.}
\begin{tabular*}{\hsize}{@{\extracolsep{\fill}}lllll@{}}
\toprule
Model & Environment & Precision & Recall & F1 score (macro) \\
\colrule
Qwen2.5:14b & few-shot 1 example & 0.52 & 0.49 & 0.51 \\
OpenAI GPT-4-turbo & few-shot 1 example  & 0.67 & 0.78 & 0.68 \\
Gemini 2.0 Flash Thinking & few-shot 1 example  & 0.65 & 0.80 & 0.77 \\
Gemini 2.0 Pro & few-shot 5 examples & 0.81 & 0.80 & 0.80 \\
\textbf{OpenAI O1} & \textbf{few-shot 5 examples} & \textbf{0.82} & \textbf{0.82} & \textbf{0.82} \\
\botrule
\label{table:abstract_level}
\end{tabular*}
\end{table}

\subsection{Performance at third task: classify strength of associations}

Our research represents the first systematic investigation into classifying the strength of associations, revealing significant challenges for LLMs and MLMs in this nuanced task, as seen in \autoref{table:strength_level}. The results underscore the difficulty of this novel problem, with models demonstrating modest performance across different experimental settings. In the few-shot scenario with a single example, Gemini 2.0 Pro and OpenAI O1 showed limited capabilities, achieving F1 scores of 0.34 and 0.38 respectively. The fine-tuned Gemini 2.0 Pro model demonstrated more balanced performance metrics, with precision and recall both approaching 0.61, suggesting a more robust approach to classifying association strengths. 

\begin{table}[h]
\caption{Evaluation results for determining the strength of associations. The finetuned version of Gemini 2.0 Pro achieves the highest performance (Precision, Recall, F1 score) outperforming few-shot approaches and ModernBERT.}
\begin{tabular*}{\hsize}{@{\extracolsep{\fill}}lllll@{}}
\toprule
Model & Environment & Precision & Recall & F1 score (macro) \\
\colrule
Gemini 2.0 Pro & few-shot 1 example  & 0.41 & 0.52 & 0.34 \\
ModernBERT & fine-tune & 0.35 & 0.35 & 0.36 \\
OpenAI O1 & few-shot 1 example  & 0.44 & 0.58 & 0.38 \\
\textbf{Gemini 2.0 Pro} & \textbf{fine-tune} & \textbf{0.60} & \textbf{0.61} & \textbf{0.60} \\

\botrule
\label{table:strength_level}
\end{tabular*}
\end{table}

\section{Conclusion and Future Work}
We benchmarked the latest NLP models, including MLMs, hybrid architectures, and cutting-edge LLMs (Gemini 2.0, OpenAI O-series, Qwen, Deepseek, Mistral), on extracting SNP-phenotype associations from the SNPPhenA corpus. We tackled the three main tasks and our results show that proprietary LLMs excel, with OpenAI O1 achieving state-of-the-art performance for non-finetuned sentence-level (F1 0.89) and abstract-level (F1 0.82) classification using few-shot learning. Classifying association strength proved challenging, though fine-tuned Gemini 2.0 Pro achieved the best result (F1 0.60), highlighting this as a complex but important area. The findings underscore the power of the latest LLMs for automating knowledge discovery in genomics, significantly outperforming MLMs, hybrid models, and the tested open-source alternatives in zero-shot and few-shot scenarios for these tasks. Despite significant advancements in open-source LLMs in recent years, a discernible performance gap persists compared to proprietary alternatives.

Our future efforts will go into improving association strength classification through refined models, enhanced fine-tuning, and more sophisticated prompting techniques. Additionally, exploring explainability methods to understand model reasoning is very  important in medical tasks because clinicians need to trust and verify the AI's suggestions before making decisions. Understanding how a model reaches a conclusion helps ensure patient safety by allowing experts to identify potential flaws, biases, or reliance on incorrect features.

\section*{Acknowledgements}
This research was partially supported by the project “Romanian Hub for Artificial Intelligence - HRIA”, Smart Growth, Digitization and Financial Instruments Program, 2021-2027, MySMIS no. 334906.








\clearpage

\end{document}